\documentclass[letterpaper, 10pt, conference]{ieeeconf}  

\IEEEoverridecommandlockouts                              

\usepackage[backend=biber,style=ieee,natbib=true]{biblatex} 
\usepackage{graphicx}
\usepackage{amsmath} 
\usepackage{amssymb}  
\usepackage{algorithm}
\usepackage[noend]{algpseudocode}
\usepackage{hyperref}
\usepackage{balance}
\hypersetup{
breaklinks=true,letterpaper=true,colorlinks,linkcolor=[RGB]{153,27,30},citecolor=[RGB]{153,27,30},urlcolor=[RGB]{153,27,30}
    }

\usepackage{amsfonts}
\usepackage{multicol}
\usepackage[normalem]{ulem}
\usepackage{booktabs}

\expandafter\def\expandafter\normalsize\expandafter{%
    \normalsize%
    \setlength\abovedisplayskip{3pt}%
    \setlength\belowdisplayskip{3pt}%
}

\usepackage{titlesec}
\titlespacing*{\section}{0pt}{4px}{4px}

\title{\LARGE \bf
MILE: Model-based Intervention Learning
}

\author{Yigit Korkmaz, Erdem B{\i}y{\i}k
\thanks{Both authors are with Thomas Lord Department of Computer Science, University of Southern California. Emails: \{ykorkmaz, biyik\}@usc.edu%
}}

\begin{document}

\maketitle
\thispagestyle{empty}
\pagestyle{empty}

\begin{abstract}

    Imitation learning techniques have been shown to be highly effective in real-world control scenarios, such as robotics. However, these approaches not only suffer from compounding error issues but also require human experts to provide complete trajectories. Although there exist interactive methods where an expert oversees the robot and intervenes if needed, these extensions usually only utilize the data collected during intervention periods and ignore the feedback signal hidden in non-intervention timesteps. In this work, we create a model to formulate how the interventions occur in such cases, and show that it is possible to learn a policy with just a handful of expert interventions. Our key insight is that it is possible to get crucial information about the quality of the current state and the optimality of the chosen action from expert feedback, regardless of the presence or the absence of intervention. We evaluate our method on various discrete and continuous simulation environments, a real-world robotic manipulation task, as well as a human subject study. Videos and the code can be found at \href{https://liralab.usc.edu/mile/}{https://liralab.usc.edu/mile}.

\end{abstract}

\section{Introduction}
\label{sec:introduction}

    Imagine training a household robot to help users place the dishes in the dishwasher. One way to do this is to use reinforcement learning (RL) that has been proven successful in several areas ranging from gaming to dialogue systems and autonomous driving \cite{kalashnikov2018scalable,levine2016end,ouyang2022training,silver2017mastering}. However, its need for lots of online interactions with the environment as well as a well-defined reward function make it unsuitable in a real-world situation like this. An alternative is to use imitation learning (IL) where an expert provides demonstrations of how to place the dishes. IL requires fewer interactions in the world than RL and does not require a reward function. A common drawback of this approach is the compounding distributional shift, which results from the accumulation of errors when deploying a learned policy \cite{ross2010efficient}: the small inaccuracies in the robot's learned policy will move it to an out-of-distribution state where the policy may fail more significantly, which may result in the robot breaking the dishes.

    Interactive learning methods try to overcome this compounding errors problem by iteratively querying the expert with system states, and fine-tuning the policy based on the expert actions \cite{ross2011reduction, kelly2019hgdagger,,kunal2019ensembledagger,hoque2022thriftydagger,hoque2023fleet,baraka2025human}. Most of these methods do not let the human intervene at will but transfer the control to the human according to some criterion \cite{kunal2019ensembledagger, hoque2022thriftydagger}. In others, the human can take over the control at any timestep \cite{kelly2019hgdagger, liu2022robot, mandlekar2020humanintheloopimitationlearningusing, spencer2020learning}.

    Going back to the running example, assume the robot had a mediocre policy in the beginning, thanks to some initial training done in the factory setup. In this interactive scheme, we would control the robot only when we think it is doing or about to do something wrong. This is clearly more convenient than operating the robot for long periods to generate full demonstrations. But what about the intervals where we did \emph{not} input any actions? Did we not provide any information to the robot? The answer is no: the fact that we \emph{chose} to not intervene means the robot's actions were already good enough in those intervals. This is what the existing interactive learning techniques are missing: even though they try to improve the robot's policy based on the states where the expert intervenes, they do not utilize any structure of how or when those interventions occur, ignoring an important feedback signal that is leaking through the states where the expert does not intervene.

    To efficiently use the information in both the states with and without human interventions, we argue one must understand and utilize the structure behind how interventions occur. To this end, we make the following contributions:
    \begin{enumerate}
        \item{We propose a novel model that is fully differentiable to formulate how and when an expert intervenes.}
        \item{We utilize this intervention model to fine-tune a weak policy and evaluate our method in various simulated and real-world environments to prove its effectiveness.}
        \item{We compare our method against the state-of-the-art baselines which utilize interventions to show its higher sample-efficiency and performance.}
    \end{enumerate}

    \begin{figure*}[t] 
        \centering
        \includegraphics[width=1.0\linewidth]{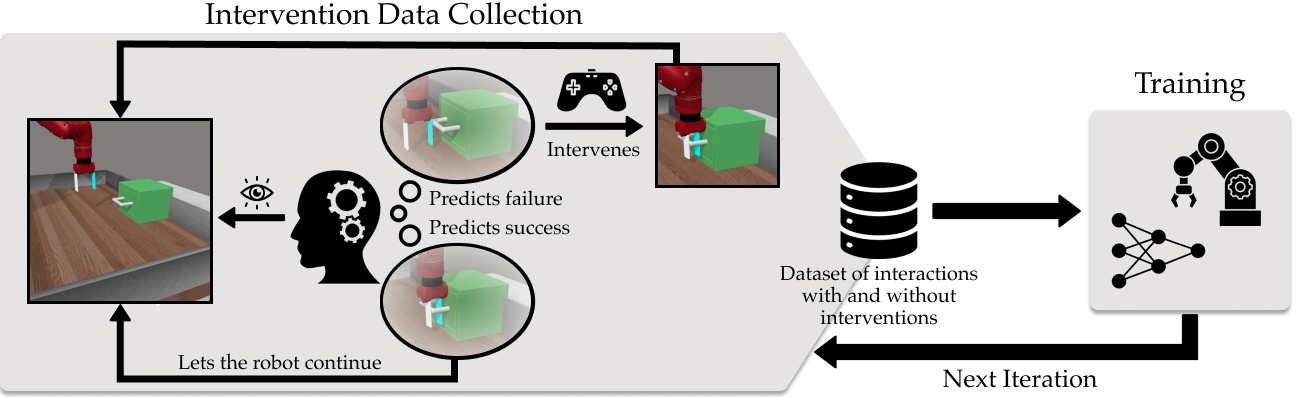}
        \vspace{-20px}
        \caption{\textbf{Overall MILE System.} A human operator oversees the robot during task execution and may decide to take over control at any timestep. The human makes the decision to intervene or not based on their prediction of the robot's potential failure, without observing the robot's action in that particular state. During the data collection phase, all interactions with the environment are recorded, both with and without interventions. The policy is then trained on this dataset using an iterative process that incorporates our novel intervention model.}
        \label{fig:front_figure}
        \vspace{-15px}
    \end{figure*}

\section{Related Work}
\label{sec:related_work}
    \textbf{Interactive Imitation Learning.} In imitation learning, expert data is used to train a policy in a supervised way \cite{celemin2022interactive, argall2009survey, ho2016general, laskey2017dart, abbeel2010autononous, pomerleau1988alvinn}. Interactive imitation learning methods try to overcome its compounding errors problem \cite{ross2010efficient, ross2011reduction} by querying the expert on the learned policy's rollouts \cite{celemin2022interactive, kelly2019hgdagger, kunal2019ensembledagger, ross2011reduction, hoque2022thriftydagger,myers2023active}. While the expert relabels either the whole trajectory of the policy with actions, or some parts of it that are automatically selected by estimating various task performance quantities \cite{hoque2022thriftydagger, ross2011reduction}, there are also works where the expert has the freedom to intervene at will \cite{kelly2019hgdagger, liu2022robot, mandlekar2020humanintheloopimitationlearningusing, chisari2021correct}. Some of these methods consider the implicit feedback coming from the states where the expert chooses to not intervene. They attempt to incorporate the information leaking from non-intervention intervals either by enforcing those state-action pairs to be constrained in the action-value cost functions \cite{spencer2020learning} or utilize weighted behavioral cloning (BC) to incorporate the signal from non-interventions, with different heuristics used for assigning weights to on-policy robot samples, and expert human interventions \cite{liu2022robot, mandlekar2020humanintheloopimitationlearningusing, chisari2021correct}.
    However, none of these algorithms uses a model to understand and learn from why the expert chooses to not intervene. In our method, we propose a model that attempts to capture how the interventions occur, and how satisfied the user is with the current action in the case of not intervening. Moreover, we investigate how we can efficiently use the information coming from that model for training. To our knowledge, this is the first work that models how human interventions happen in robotics while utilizing it for better policy learning.

    \noindent\textbf{Reinforcement Learning with Expert Data.} Many studies focus on integrating expert demonstrations into RL. This is often achieved by populating the replay buffer to warm-start the initial policy and guide its exploration in a favorable direction \cite{hester2018deep, ball2023efficient, nair2018overcoming, xie2021policy}. There are also approaches that incorporate interventions into RL. One approach introduces an extra cost term in the learning objective to minimize interventions \cite{li2022efficient}. Another recent method relabels the rewards based on interventions, assigning negative rewards at intervention states, while giving a zero reward to all other states \cite{luo2023rlif}. However, most of these methods assume access to some type of reward information, either through explicit reward functions or through Q-functions of the task. Our method does not require these assumptions. Not to mention, these methods still require a lot of online interactions with the environment and RL algorithms that can handle both on-policy and off-policy data coming from the intervening agent.

    \noindent\textbf{Modeling Interventions.} There have been efforts to provide formal definitions and qualitative analyses of interventions in the human-robot interaction research \cite{wang2025effects}, along with developing metrics to evaluate operator workload \cite{scholtz2003theory, steinfeld2006common}. Within computational modeling, various studies have addressed the modeling of trust in human-robot collaborative settings, which include both binary and continuous measures \cite{rodriguez2023review}. There are also works exploring the temporal dynamics of trust conditioned on task performance \cite{guo2021modeling, chen2018planning}. One such work models trust as a latent variable in a partially observable Markov decision process, utilizing this model to maximize task performance \cite{chen2018planning}. To our knowledge, no prior work has developed a similar computational model to explain the occurrence of interventions in human-robot teams.

\section{Problem Definition}
\label{sec:problem_definition}

    We formulate the problem as a discrete time Markov decision process (MDP) with the standard $\langle S,A,\rho,f,R,T,\gamma\rangle$ notation. The robot does not know the reward function $R$ or the transition dynamics $f$ of the environment. It starts with an initial policy $\pi_\theta: S \rightarrow A$ parameterized by $\theta$.
    
    In addition, a potentially noisy expert human can intervene (take over the control) with any action $a_h\in A$ at any timestep. However, in contrast to some of the existing works \cite{saunders2018trial}, the human has to make the decision about intervening or not without observing the robot's action $a_r \in A$ in that particular state. This is a more viable setting in robotics where it is not realistic to expect the robot to check each and every action with the human.
    
    We let the robot and the human interact in this setup, and we record $(s,a_r,a_h,s')$ from every timestep, with the possibility that $a_h$ is undefined if the human did not intervene in that timestep. Our objective is to find a policy \(\pi_{\theta^*}\) that maximizes the expected cumulative discounted reward:
    \begin{align*}
    \theta^* = \arg\max_\theta \mathbb{E}_{\tau\sim\pi_\theta}\left[\sum_{t=0}^{T-1}\gamma^t R(s_t,a_t)\right]
    \end{align*}
    where $s_0\sim\rho(\cdot)$, and $(s_t,a_t)$ pairs are sampled according to the policy $\pi_\theta$ and transition function $f$.
    
    We assume the human interacts with the robot for $N$ iterations, each of which includes $k$ episodes. After each iteration, we train the robot policy using the data collected so far. In the next section, we describe how we do this training.

\section{Model-based Intervention Learning (MILE)}
\label{sec:method}

    Our method to solve this problem is based on a computational model of when and how the human may decide to intervene the robot's operation. We will introduce this intervention model in the first subsection. Subsequently, we will leverage this model to create a framework where we update the robot's policy $\pi_\theta$ based on its interactions with the human and the environment.
    
    \begin{figure*}[t] 
        \centering
        \includegraphics[width=\linewidth]{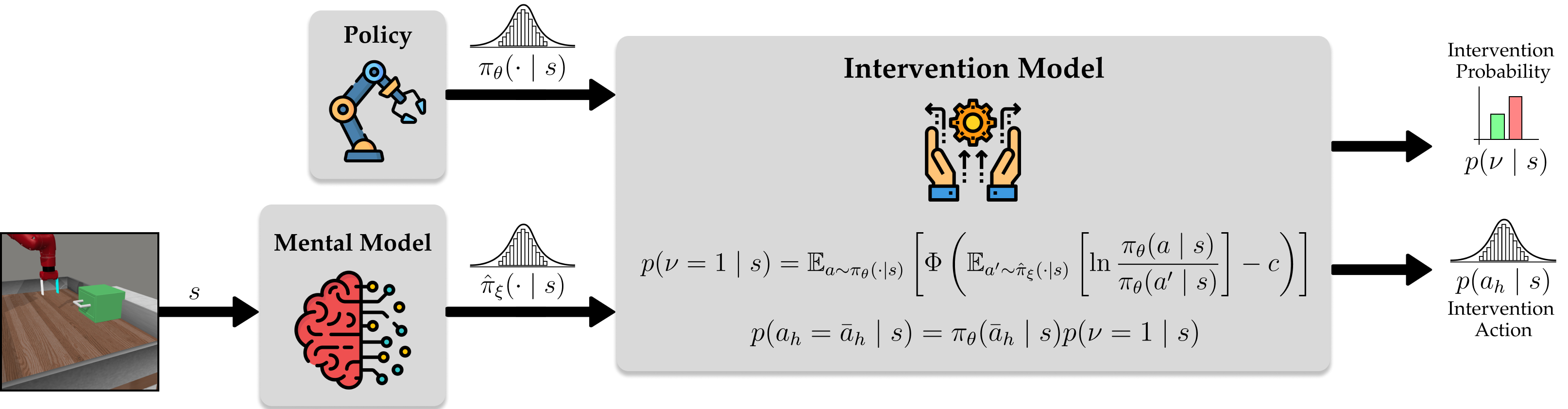}
        \vspace{-20px}
        \caption{\textbf{Framework for learning from interventions.} Starting from an inital policy $\pi_\theta$, we jointly train the mental model, and the policy using our intervention model.}
        \label{fig:pipeline}
        \vspace{-15px}
    \end{figure*}
    
    \textbf{Intervention Model.} Consider our running example: would you intervene the robot's operation when it is quickly moving toward a glass jar that stands at the edge of the countertop? How about if the jar was made of plastic? This example highlights the important factors that affect when humans intervene the robot. Firstly, consider the case of glass jar. The human is likely to intervene because they predict the robot may hit and break the jar. So the human's belief about what the robot will do, i.e., their mental model of the robot, affects their decision to intervene, as this was previously shown in human-robot interaction research \cite{chen2018planning}.

    Next, let's compare the cases with glass and plastic jars. The human is less likely to intervene in the latter, because even if the robot takes bad actions, the outcome is not going to be as bad as in the case of glass jar. This points out that the human's interventions also depend on how good/bad the potential outcomes are.
        
    Considering these factors, we propose an intervention model based on the probit model from discrete decision theory \cite{bliss1934probit}.
    For the ease of the presentation, we will start with discrete action spaces and then extend our formulation to continuous domains.
    
    Let $\nu$ be a binary random variable that indicates whether the human intervenes ($\nu=1$) or not ($\nu=0$), and $\bar{a}_h$ denote the nominal human action, i.e., the action human would take provided they decide to intervene. Mathematically, $a_h=\bar{a}_h$ if and only if $\nu=1$. Otherwise, $a_h$ is not defined in that state. Finally, let $\hat{\pi}$ denote the human's mental model of the robot, i.e., what the human believes the robot will do in a given state. This prediction is needed as the human has to intervene before seeing the robot’s action. While some learning from intervention works assumes the human first sees the system’s proposed action \cite{saunders2018trial}, ours is a more realistic setting, especially in robotics where conveying an action without taking it is both difficult and time-consuming.

    We start with modeling the probability that the human will intervene at a given state $s$. We use $\sigma$ for the softmax operation and $\Phi$ for the cdf of a standard normal distribution.
    \begin{align}\label{eq:intervention}
    p(\nu=1\mid s) &= \sum_{a\in A}p(\bar{a}_h=a,\nu=1\mid s) \nonumber\\
                   &= \sum_{a\in A}p(\bar{a}_h=a\mid s)p(\nu=1\mid \bar{a}_h=a,s)
    \end{align}
    Here, the first term inside the summation is just the probability that the human would take action $a$ at state $s$. Since we assume the human is a (noisy) expert, we use a Boltzmann policy under the true reward function of the environment to model this probability, similar to the prior work \cite{dragan2013policy, javdani2015shared, kwon2020when, ziebart2008Maximum, laidlaw2022boltzmann}:
    \begin{align*}
        p(\bar{a}_h\!=\!a\!\mid\! s) \!=\! \pi_h(a \!\mid\! s) \!:=\! \sigma(Q(s,a)) \!=\! \frac{\exp(Q(s,a))}{\sum_{a'\in A}\! \exp(Q(s,a'))}
    \end{align*}
    The second term in Eq.~\eqref{eq:intervention} denotes the probability that the human will intervene conditioned on the state and their nominal action. This is where we bring the probit model into play: the human will intervene only if their nominal action is considerably better than what they expect the robot to do:
    \begin{align*}
        p(\nu\!=\!1\mid \bar{a}_h\!=\!a,s) &=\Phi\Big(Q(s,a) \!-\! \mathbb{E}_{a'\sim \hat{\pi}(\cdot\mid s)}[Q(s,a')] \!-\! c\Big)\:,
    \end{align*}
    where $c$ is a scalar hyperparameter. In this probit-based model, the human is more likely to intervene if the value of their nominal action is much higher than the expected value of the robot's action. The scalar $c$ depends on the effort the human needs to put to intervene the robot. If it is difficult to intervene, $c$ is going to be high and the human will only intervene if the difference in action values is extremely high. On the other extreme, if interventions are free, then $c$ will be low and the human may continually intervene.

    Now that we modeled the probability of human interventions, we continue with \emph{how} they intervene when they do:
    \begin{align*}
    p(a_h=\bar{a}_h\mid s) &= p(a_h=\bar{a}_h\mid \nu=0,s)p(\nu=0\mid s) \nonumber\\
                           &\quad + p(a_h=\bar{a}_h\mid\nu=1,s)p(\nu=1\mid s) \nonumber\\
                           &= \sigma(Q(s,\bar{a}_h))p(\nu=1\mid s)\:,
    \end{align*}
    where $p(a_h=\bar{a}_h\mid \nu=0,s)=0$ by the definition of $a_h$.

    We note the following relation due to the Boltzmann policy formulation we use for the human:
    \begin{align*}
    \ln \pi_h(a \mid s) - \ln \pi_h(a' \mid s) &= \ln\left(\frac{\exp Q(s,a)}{\sum_{a'' \in A} \exp Q(s,a'')}\right) \nonumber\\
    &\quad - \ln\left(\frac{\exp Q(s,a')}{\sum_{a'' \in A} \exp Q(s,a'')}\right) \nonumber\\
    &= Q(s,a) - Q(s,a')\:.
    \end{align*}
    This allows us to rewrite $p(\nu=1\mid s)$ and $p(a_h=\bar{a}_h\mid s)$ only dependent on the policy and not the $Q$-function:
    \begin{align}
        &p(\nu=1\mid s) \label{eq:when_final}\\
        &=\!\sum_{a\in A}\!\pi_h(a \!\mid\! s)\Phi\left(\mathbb{E}_{a'\sim \hat{\pi}(\cdot\mid s)}[\ln \pi_h(a \mid s) \!-\! \ln\pi_h(a' \mid s)]\!-\!c\right) \nonumber\\
        &=\!\mathbb{E}_{a\sim\pi_h(\cdot \mid s)}\left[\Phi\left(\mathbb{E}_{a'\sim \hat{\pi}(\cdot \mid s)}[\ln \pi_h(a \mid s) \!-\! \ln\pi_h(a' \mid s)]\!-\!c\right)\right] \nonumber\\
        &p(a_h=\bar{a}_h\mid s) = \pi_h(\bar{a}_h\mid s)p(\nu=1\mid s)\label{eq:how_final}
    \end{align}
    This change of variables allows us to readily use this intervention model in continuous domains. The main difference is that, in the discrete case, $a_h$ has a categorical distribution over the action space as well as the no-intervention ($\nu=0$). In continuous case, $\nu$ acts like a gate controlling which continuous policy ($\pi_h$ or $\pi_\theta$) will be active. We will now present how we use this intervention model to learn from humans' intervention feedback.

    \textbf{Learning from Interventions.} The robot misses information about two critical components in the intervention model: $\hat{\pi}$ that models what the human thinks the robot will do, and the human's policy $\pi_h$. In our learning algorithm, we model both of these policies with neural networks, $\hat{\pi}_\xi$ and $\pi_\theta$, respectively (see ``mental model'' and ``policy'' in Fig.~\ref{fig:pipeline}).
    Since the intervention model is differentiable, we conveniently utilize the gradients coming from it to jointly train these networks using the dataset of $(s,a_r,a_h,s')$ tuples.
    During inference time, we only employ the trained policy $\pi_{\theta}$.
    
    In discrete domains, we minimize the cross entropy loss between the ground truth actions ($a_h$ if $\nu=1$ and $a_r$ otherwise) in the dataset and estimated final action probabilities, where no-intervention ($\nu=0$) is an extra class in the action space. Mathematically, the loss function is:
    \begin{align*}
    J(\theta, \xi) = \text{CE}(\hat{\mathbf{P}}(\mathbf{S}; \theta, \xi), \mathbf{a_h}) \!=\! -\frac{1}{N} \sum_{i=1}^N \log(\hat{p}_{i, a_h^i}(\mathbf{s}_i; \theta , \xi))\:,
    \end{align*}
    where $\mathbf{S}$ is the set of states in the dataset, $\hat{\mathbf{P}}(\mathbf{S}; \theta, \xi)$ is the matrix of probabilities with each row $\hat{\mathbf{p}}_{i}(\mathbf{s}_i; \theta, \xi)$ being the predicted probability vector for sample $i$, $\mathbf{a_h}$ is the vector of true intervention action labels including no-intervention, and $\hat{p}_{i, a_h^i }(\mathbf{s}_i; \theta, \xi)$ is the predicted probability for the true class of sample $i$.
    
    In continuous domains, we use a combination of discrete and continuous loss functions as we want our model to learn both when and how the human intervenes. In order to accomplish the former, we use binary cross entropy loss between ground truth intervention signals $\nu$ and the estimated probability of intervention from Eq.~\eqref{eq:when_final}, where $\pi_{\theta}$ replaces $\pi_h$ and $\hat{\pi}_{\xi}$ replaces $\hat{\pi}$:
    \begin{align*}
    &J_{1}(\theta, \xi) = \text{BCE}(\hat{\pmb{\nu}}(\mathbf{S}; \theta, \xi), \pmb{\nu}) \nonumber\\
    &=\! -\frac{1}{N} \sum_{i=1}^N \!\left[ \nu_i \log\left(\hat{\nu}(\mathbf{s}_i; \theta, \xi)\right) \!+\! (1 \!-\! \nu_i) \log\left(1 \!-\! \hat{\nu}(\mathbf{s}_i; \theta, \xi)\right) \right],
    \end{align*}
    where $\hat{\pmb{\nu}}(\mathbf{S}; \theta, \xi)$ is the predicted probability of whether an intervention will happen. For the latter, we minimize the negative log-likelihood of the human actions $a_h$ (when $\nu=1$) under the estimated action distribution in that state:
    \begin{align}
    J_{2}(\theta) = \text{NLL}(\pi_{\theta}(\mathbf{a_h} \mid \mathbf{S})) = -\frac{1}{N} \sum_{i=1}^N \log(\pi_{\theta}(a_h^i \mid \mathbf{s}_i))
    \end{align}
    $J_1$ (intervention loss) updates both the policy and the mental model, while $J_2$ (policy loss) updates only the policy $\pi_{\theta}$, both terms contributing to policy learning:
    \begin{align}
    J(\theta, \xi) = \lambda J_{1}(\theta, \xi) + (1-\lambda) J_{2}(\theta)
    \label{eq:total_loss}
    \end{align}
    In our experiments, we set $\lambda = 0.5$. We show the complete pipeline in Fig.~\ref{fig:pipeline}, and the pseudocode on the project website.

\section{Simulation Experiments}\label{sec:sim}
    
    \begin{figure*}[t] 
        \centering
        \includegraphics[width=1.0\linewidth]{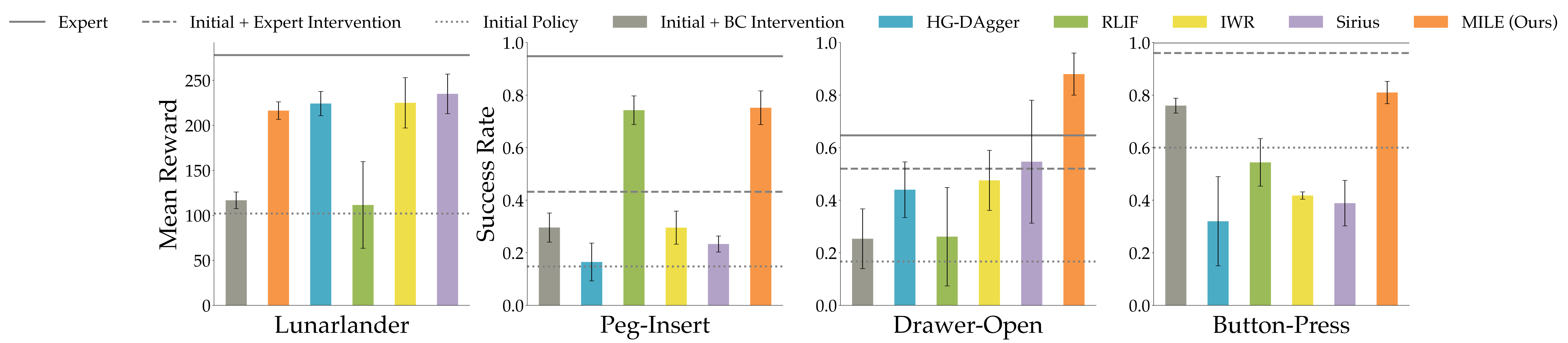}
        \vspace{-20px}
        \caption{\textbf{Success rates for single iteration training (mean$\pm$std).} MILE was trained with $N=1$ iteration and $k=15$ episodes. To make it a fair comparison, we train the ohter baselines until they have the same number of interventions in their datasets as ours.}
        \vspace{-20px}
        \label{fig:main_results}
    \end{figure*}

    \textbf{Experiment Setup.} We tested our method on four different simulation tasks, one with a discrete and three with continuous action spaces. The discrete action space environment is Lunarlander from Gymnasium \cite{towers_gymnasium_2023}.
    The remaining three tasks—Drawer-Open, Peg-Insertion, and Button-Press—are from the Metaworld suite \cite{yu2019meta} where the primary goal is to control a 6 DoF robot arm with a gripper to accomplish various tasks that require precise joint control and are prone to distributional shifts and local minima issues during RL optimization. For the Metaworld experiments, we use true world states as observations. To retain temporal information, we concatenate the states of the previous three timesteps with the current timestep. The action space is composed of 4 DoF: the Cartesian positions of the end effector, which always points down, and +1 dimension for the gripper. Since our model does not rely on any assumptions about the environment's reward structure, and our primary comparisons are with other interactive learning methods, we follow the prior work \cite{sontakke2023roboclip} to evaluate the methods using success rates. Lunarlander is the only exception where we used the environment rewards due to the simplicity of the task.
    
    We use simulated humans to mimic human interventions in our simulation experiments. The goal of these experiments is to scale up the tasks and to show that it is possible to fine-tune an initial policy in a sample-efficient way using our framework. The data collection phase involves two distinct policies: a suboptimal agent for generating rollouts and an expert agent for interventions guided by our intervention model. In place of the simulated human's mental model of the robot ($\pi_\zeta$), we train a BC policy on the rollouts of the suboptimal policy. During data collection, we rollout the suboptimal policy and intervene with the simulated human. In other words, the policy $\pi_h$ of the human in the intervention model is replaced by $\pi^*$ during the data collection using simulated humans. We see that these simulated interventions increase the success rate by comparing the dashed and the dotted lines in Fig.~\ref{fig:main_results}. 
    During training, we jointly train the suboptimal agent and the mental model as we described in Section~\ref{sec:method}. At test time, we only use the trained policy and discard the mental model. 

    \textbf{Baselines.} We compare our methods against (i) BC intervention where the model estimates human's interventions with only a neural network, (ii) HG-DAgger \cite{kelly2019hgdagger}, a state-of-the-art algorithm that iteratively finetunes the policy based on the interventions, (iii) RLIF \cite{luo2023rlif}, a recent work which models the interactive imitation learning as an RL problem and uses interventions as the reward signals to train the policy in an online fashion, (iv) IWR \cite{mandlekar2020humanintheloopimitationlearningusing} and (v) Sirius \cite{liu2022robot}, two methods utilizing weighted BC with different heuristics to learn from both intervention and non-intervention timesteps. We start with the same initial suboptimal policy for all methods, and assume there is no access to any expert dataset, apart from the interventions provided.

    To be consistent with the realistic settings, we keep the intervention ratio in the rollouts to be less than 30$\%$ by tuning the hyperparameter $c$.
    For our initial results, we train our method with $N=1$ iteration and $k=15$ episodes. For fairness, we train the other baselines until they have the same number of interventions in their datasets as ours.

    \textbf{Results.} We present the results in Fig.~\ref{fig:main_results}. Our method achieves the best results across all environments (only tied with HG-DAgger, IWR and Sirius in Lunarlander, and with RLIF in Peg-Insert), proving its sample-efficiency. Due to the small amount of data which mostly consist of states with no interventions, IL methods suffer from overfitting as well as compounding errors. On the other hand, RLIF fails to produce reliable results, possibly due to the number of samples being too few to learn a successful policy.

\begin{figure}[b]
        \centering
        \vspace{-15px}
        \includegraphics[width=1.0\linewidth]{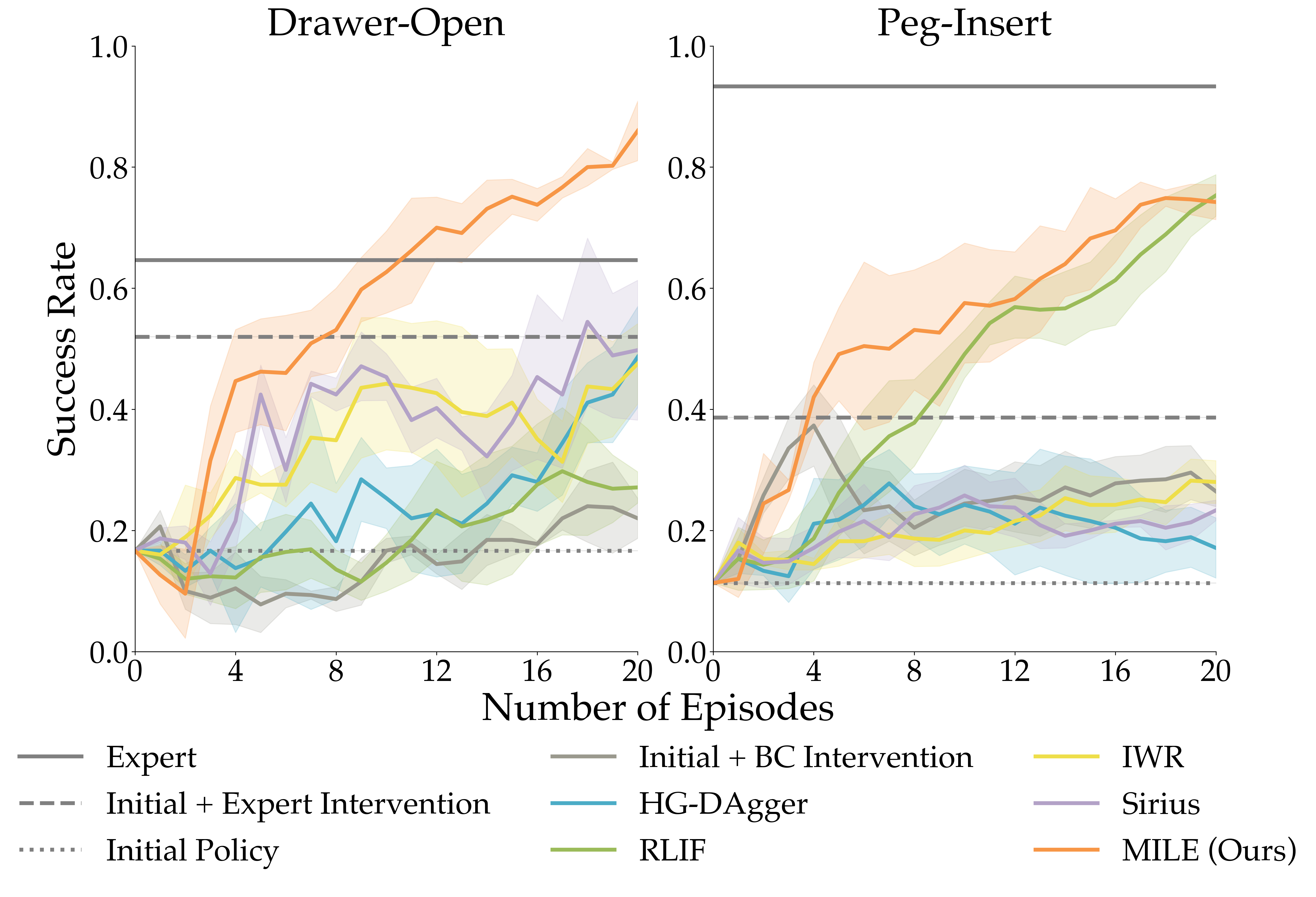}
        \vspace{-20px}
        \caption{\textbf{Success rates for iterative training (mean$\pm$s.e.).}}
        \label{fig:online_results}
    \end{figure}

    To align more with the online nature of the baselines, we also perform iterative training with $N=20$ iterations and $k=1$ in Drawer-Open and Peg-Insert environments, as Button-Press was a relatively easier task. We run HG-DAgger, IWR, and Sirius in the same setting. For RLIF, we set the number of samples in its replay buffer approximately the same as MILE and other methods. The results are shown in Fig.~\ref{fig:online_results}. We observe that our method significantly outperforms the baselines. In just $10$ iterations, it is able to achieve the success rate of the expert policy which provided the interventions, following with an even greater performance while the baselines mostly fall short.

\begin{figure}[bt]
        \centering
        \includegraphics[width=1.0\linewidth]{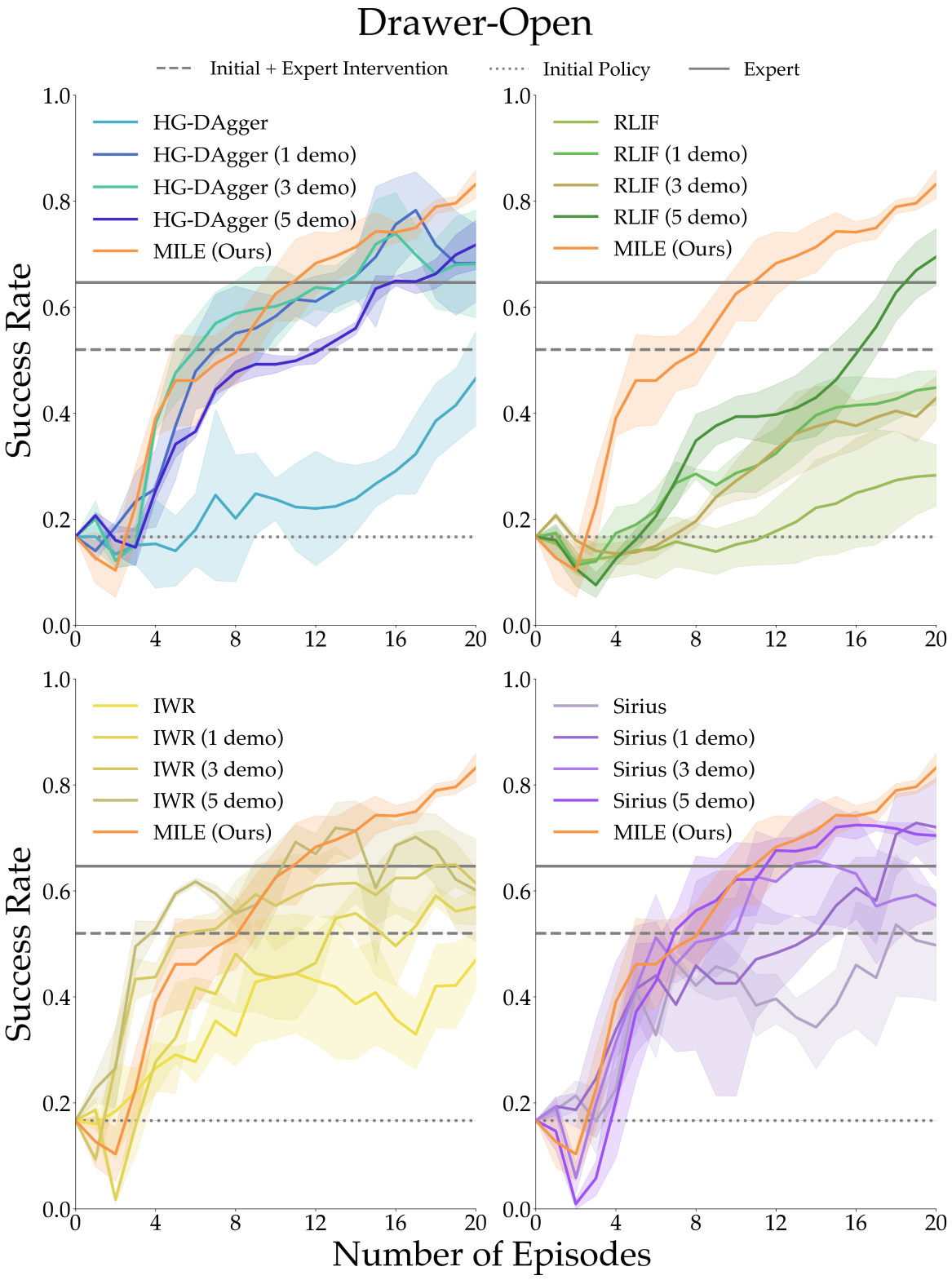}
        \vspace{-20px}
        \caption{\textbf{Success rates for the demo ablation study (mean$\pm$s.e.).}}
        \vspace{-20px}
        \label{fig:demo_ablation}
    \end{figure}

    \textbf{Offline Demonstration Ablation.} All of the baselines we compare against normally assume access to a set of offline demonstrations. Although MILE does not require such access, we run an ablation study in Drawer-Open, comparing our method against the baselines when they have access to expert demonstrations.
    Results are shown in Fig.~\ref{fig:demo_ablation}. Our method performs better than RLIF even when it has access to 5 expert demonstrations. 
    IL-based methods' performance increases with the number of demonstrations in the dataset and becomes competitive with MILE but plateaus around the success rate of the expert agent. We believe this is due to overfitting and compounding errors. In both IWR and Sirius, robot actions are used as direct supervision signals in cases of non-intervention, albeit with different weights. In contrast, MILE uses the computational model for human interventions that we developed to construct probability distributions over the action space which are then used to train the models, thereby increasing the information bandwidth of the human feedback by computationally modeling them.

    Overall, simulation results show our model for interventions improves learning, and increases adaptability, especially in low data regime where other interactive learning methods fail. With a handful of intervention trajectories, we are able to achieve near-optimal success levels in various tasks.

\section{Real Robot Experiment}
\label{sec:real_robot}

\begin{figure}[t] 
        \centering
        \includegraphics[width=\linewidth]{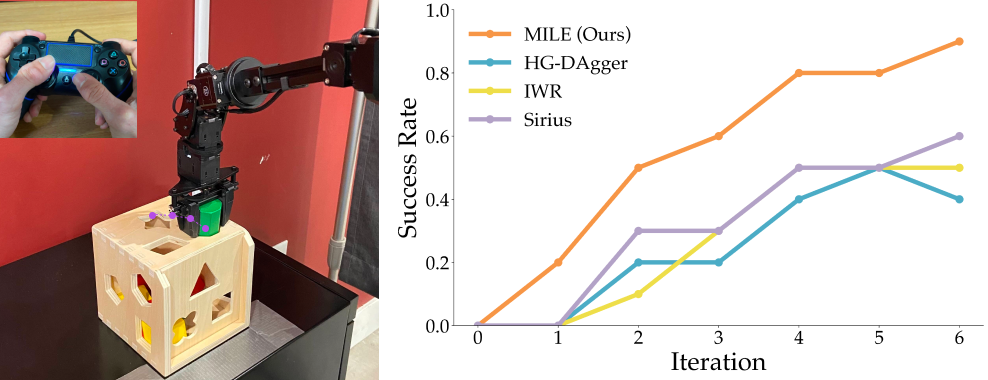}
        \vspace{-20px}
        \caption{\textbf{Real robot experiment setup and Success rates for MILE and other IL-based methods.} All methods were trained for 6 iterations with collecting 3 trajectories each iteration.}
        \vspace{-15px}
        \label{fig:real_world}
    \end{figure}

    In order to show the effectiveness of our method in real world settings, we designed an experiment using a WidowX 6-DoF robot arm and with a real human. The task is to put the octagonal block into the wooden box through the correct hole as shown in Fig.~\ref{fig:real_world}. The sensitivity of this task is very high, as there exists a very small margin between the block and the hole, which makes it a suitable candidate when the small interventions provided by the humans can have substantial effect on the robot's success. We used image observations along with the robot's end effector position.
    
    The robot is initialized with a mediocre policy that is unable to complete the task. During the execution, the human can take over the controls of the robot at any time via a PS3 controller. Similarly, they can also give the controls back to the robot. We run the experiment with $N=6$ iterations, with $k=3$ intervention trajectories collected in each iteration. 

    We compare our method against IL-based methods, which have been somewhat more successful than RLIF in the simulation studies. All methods use the same data collection interface. Similar to the simulation experiments, we do not assume access to any offline expert demonstrations and only use the intervention trajectories. The results are shown in Fig.~\ref{fig:real_world}. Our method is able to achieve 80\% success rate just after 4 iterations, while other baselines struggle to improve the policy. This result showcases that our method is applicable to real world settings as well.

\section{Human Subject Study on Real Robot}
\label{sec:user_study}

    Finally, we conducted a user study to analyze how accurately our model estimates the humans' interventions, as the success of our method relies on the success of the intervention model capturing when and how humans intervene the robot. The study is approved by the IRB office of the University of Southern California. We used the same task setting and experiment setup as in the previous section.

    \textbf{Procedure.} We recruited 10 participants (3 female, 7 male, ages 20-28). The study started with a training period where subjects had a chance to get familiar with the robot controls. 

    We had 2 experiment settings. At the beginning of each experiment, the subject was first shown 3 trajectories from the initial policy to make them aware of the robot's capabilities. In the first experiment, the user was asked to collect 5 intervention trajectories ($N=1$, $k=5$), where they can take over the initial policy at any time. These trajectories were then used to fine-tune the initial policy using our method, and the final policy was shown to the user. We asked the users to rate the initial and the final policies.

    The second experiment differed from the first by only using $N=5$ and $k=2$.
    The users could intervene and take control at any time during data collection: we did not provide any instructions on how often to intervene. We showed the finetuned policy at the end of each iteration and asked the user to rate its performance. At the end of the study, we did a post-study survey to collect participants' feedback on the method's effectiveness, improvement and the their satisfaction on the final result.

\begin{figure}[ht!]
        \centering
        \vspace{-5px}
        \includegraphics[width=1.0\linewidth]{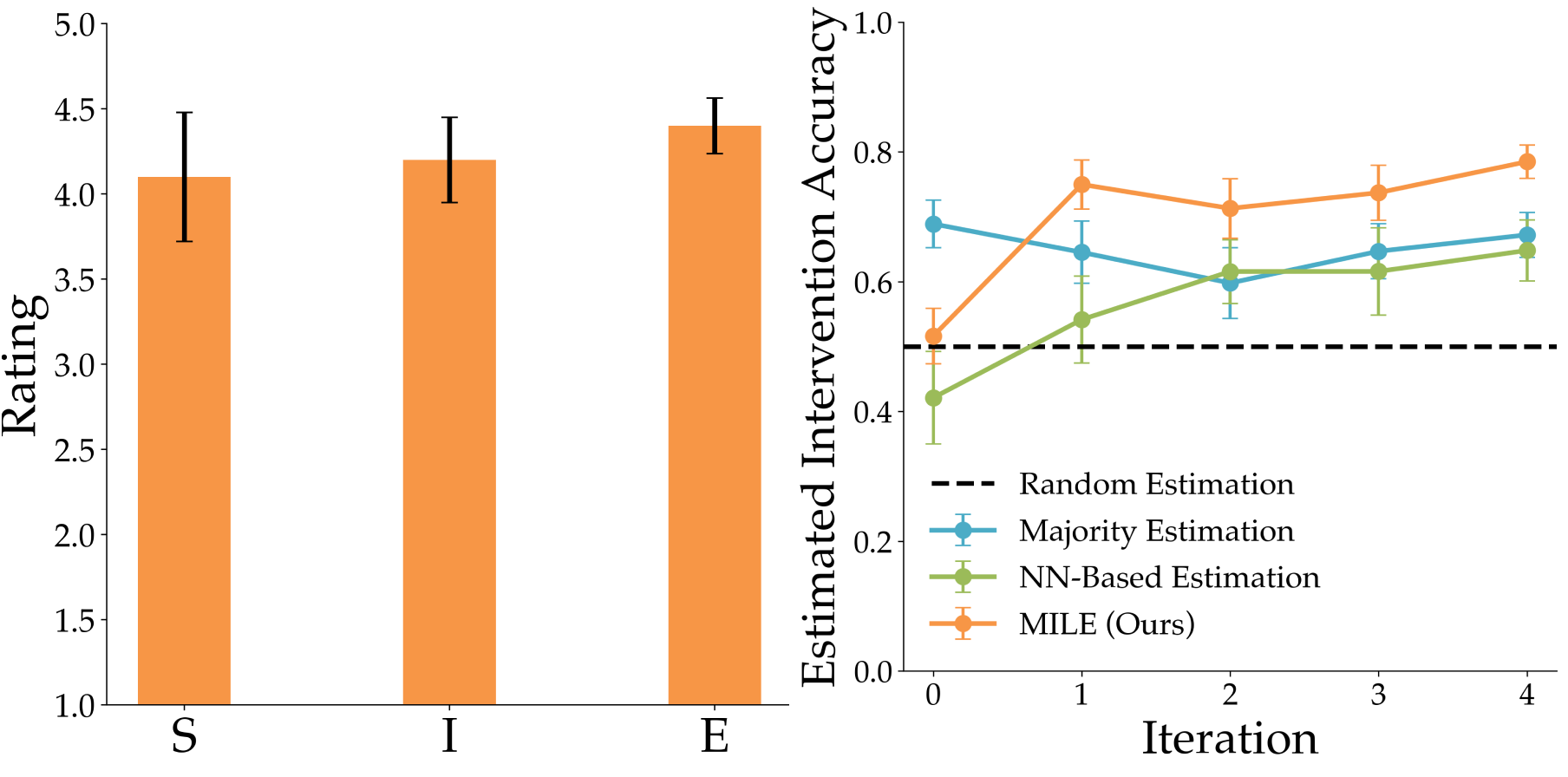}
        \vspace{-20px}
        \caption{\textbf{User Study Results (mean$\pm$s.e.).} Legend: S=Satisfaction, I=Improvement, E=Effectiveness}
        \vspace{-12px}
        \label{fig:user_study}
    \end{figure}

    \textbf{Results.} We compared our intervention model with the majority estimator that always predicts no intervention (the intervention rate of humans is around 30-40\% throughout the procedure) and a neural network based estimator trained with the data from previous iterations to classify interventions in the current iteration. As shown in Fig.~\ref{fig:user_study}, our model outperforms both methods in predicting when an intervention will occur. Additionally, most users rated our model's effectiveness and adaptation speed as adequate, demonstrating it successfully mirrors real human interventions.

\section{Conclusion}
\label{sec:conclusion}

    \textbf{Summary.} In this work, we presented MILE, a novel way of integrating interventions in policy learning. We developed a fully-differentiable intervention model to estimate when and how the interventions occur, and a policy training framework that uses the intervention model.

    \noindent\textbf{Limitations and Future Work.} Our intervention model considers only the current state to estimate interventions. This does not fully capture the temporal analysis humans make before intervening, e.g., waiting for robot to fail in a recoverable way before taking control. Future work should explore ways of integrating this temporal aspect of intervention estimation, possibly with maintaining a belief over robot actions and using Bayesian updates. Another limitation is our assumption that humans have an accurate mental model of the robot policy. Investigating how incorrect mental models affect the performance is an important direction for future research. Recently, some works showed that the interventions can be used for data generation and allow better sim-to-real generalization \cite{jiang2024transic, hoque2024intervengen}. Testing our method's capabilities in these domains and exploring whether using a structure behind interventions enables further improvements in generalization are interesting future directions. Finally, learning from natural language-based corrections \cite{yang2024trajectory,shi2024yell} is a recent exciting research area that may benefit from our intervention model.









\balance
\printbibliography

\clearpage
\onecolumn
\appendices  
\section*{Supplementary Material}

\subsection{Pseudocode of the Algorithm}
    \vspace{-5px}
    \begin{algorithm}
    \algrenewcommand\algorithmicindent{0.75em}%
    \small
    \caption{MILE: Model-based Intervention Learning in Iterative Setting}
    \label{algo:iterative_pseudocode}
    \noindent
    \begin{minipage}[c]{0.45\linewidth}
    \begin{algorithmic}
    \vspace{-32px}
    \State \textbf{Notations} \\
    $N$: maximum deployment rounds, $k$: number of rollout episodes in each deployment round, \\
    $l$: number of batches, $b$: batch size,\\
    $m$: number of epochs in each learning round, \\
    $\alpha$: learning rate, \\ 
    $\pi^{\theta}_{1}$: initial policy, $\hat{\pi}^{\xi}_{1}$: initial mental model
    
    \vspace{0.3cm}
    \For {$i \gets 1$ to $N$}
        \State $D^{i+1} \gets \text{DEPLOYMENT}(\pi^{\theta}_i, D^i)$
        \State $\pi^{\theta}_{i+1}, \hat{\pi}^{\xi}_{i+1} \gets \text{LEARNING}(\pi^{\theta}_i, \hat{\pi}^{\xi}_{i}, D^i)$
    \EndFor
    
    \end{algorithmic}

    \end{minipage}
    \begin{minipage}[c]{0.55\linewidth}
    \begin{algorithmic}
    \Function{DEPLOYMENT}{$\pi_\theta, D$}
        \State Collect rollouts w/ interventions $\tau_1, \ldots, \tau_k$
        \State $D' \gets D \cup \{\tau_1, \ldots, \tau_k\}$
        \State \Return $D'$
    \EndFunction
    \vspace{0.1cm}
    \Function{LEARNING}{$\pi_\theta, \hat{\pi}_{\xi}, D$}
        \For {$m$ epochs \textbf{and} $l$ batches each}
                \State Get the next mini-batch $\left(s^i, a_r^i, a_h^i, \nu^i\right)_{i=1}^b \sim D$
                \State Compute $\hat{\nu}(s^i; \theta, \xi) \!=\! p(\nu^i\!=\!1\!\mid\! s^i)$ based on Eq. (7)
                \State Compute $J(\theta, \xi)$ based on Eq. (12)
                \State Run in parallel:
                \State $\quad \theta \gets \theta - \alpha \nabla_\theta J(\theta, \xi)$
                \State $\quad \xi \gets \xi - \alpha \nabla_\xi J(\theta,\xi)$
        \EndFor
        \State \Return $\pi_\theta$, $\hat{\pi}_{\xi}$
    \EndFunction
    
    \end{algorithmic}
    \end{minipage}
    \end{algorithm}
    \vspace{-5px}

\subsection{Simulation Experiment Details}
Depending on the action space, we trained suboptimal initial policies and various levels of expert policies for simulated humans using SAC or DQN. We train a different policy for each task. For each task, the same expert is used to intervene across all methods. To generate the mental models of simulated humans, we collected 100 rollouts of the initial policy and trained a BC agent on them. In all result plots, we display the mean and standard error over 3 seeds for each method.

Regarding the observation space, we use true world states, i.e. low-level states of the robot and the task-related object. This includes robot joint positions and the positions and orientations of relevant objects. Our action space consists of the change in Cartesian coordinates of the robot end effector and the gripper state. Each episode has a maximum of 1000 timesteps, with early termination upon success.

To compare our method with RLIF, we used RLPD as its backbone algorithm in the domains with continuous action spaces as it was done in the original paper. For domains with discrete action spaces, we used DQN as the backbone of RLIF. We initialized the policy networks for all methods as the clones of the initial policy $\pi_\theta$. For the offline demonstration ablation, we also initialized RLIF's replay buffer with those demonstrations.

\subsection{Real Robot Experiment Details}
In this experiment, we use image observations (captured from a USB webcam) along with the robot's end effector position. We keep the same action space as in the simulation experiments for real-world settings. The robot begins with the octagonal block in its gripper. The initial BC policy is trained using 120 human-collected trajectories, gathered with a Meta Quest 2 headset. At the beginning of real-world experiments, we show 3 trajectories to the users to make them aware of the robot's capabilities. We also warm-start the initial mental model using these same demonstrations. This process requires no supervision from the human and minimal effort since the human only needs to observe the robot.

\subsection{Hyperparameters}
We used a Multi-Layer Perceptron (MLP) with hidden dimensions of 256 in both the MetaWorld and real-robot experiments for all methods. In order to get image embeddings in real-robot experiment, we used a pretrained R3M model with ResNet50 architecture. To retain temporal information, we concatenate the states of the previous three timesteps with the current timestep. For the LunarLander experiment, we used an MLP with hidden dimensions of 64 for all methods. The network outputs an action for the current step.

The log-probabilities of the actions vary in scale between the policies used in different tasks. This necessitates adjusting the standard deviation of the normal distribution whose cumulative distribution function (CDF) is used in Equation 2, along with task-specific cost terms (parameter $c$), to calculate smoother rather than skewed intervention probabilities.
\begin{table}[htb]
\centering
\caption{Hyperparameters across different simulation and real-world tasks.}
\resizebox{\linewidth}{!}{
\large
\begin{tabular}{lccccc@{}}
\toprule
          & \textbf{LunarLander} & \textbf{Button-Press}  & \textbf{Drawer-Open} & \textbf{Peg-Insert} & \textbf{WidowX Peg Insertion}  \\ \midrule
    \multicolumn{1}{@{}l}{\textbf{MILE (Ours)}}\\
    Learning Rate & 1e-5 & 1e-4 & 5e-4 & 1e-4 & 1e-5\\
    Mental Model Hidden Dims  & (64, 64) & (256, 256) & (256, 256) & (256, 256) & (256,256)\\
    Dataset size in Offline Experiment & 5 Trajectories & 15 Trajectories & 15 Trajectories & 15 Trajectories & - \\
    Number of Iterations & - & - & 20 & 20 & 6 \\
    Episodes Per Iteration & - & - & 1 & 1 & 3 \\
    Training Epochs per Iteration & - & - & 300 & 300 & 500 \\
    Intervention CDF c & 3 & 150 & 60 & 75 & 70\\
    Intervention CDF $\sigma$ & 1 & 200 & 75 & 175 & 100\\
  \midrule
    \multicolumn{1}{@{}l}{\textbf{HG-DAgger}}\\
    Learning Rate & 5e-6 & 5e-5 & 1e-4 & 5e-6 & 1e-5\\
    Number of Iterations & 5 & 5 & 5/20 & 5/20 & 6\\
    Episodes Per Iteration & 1 & 3 & 3/1 & 3/1 & 3\\
    Training Epochs per Iteration & 400 & 1000 & 1000/300 & 1000/300 & 500 \\
  \midrule
    \multicolumn{1}{@{}l}{\textbf{RLIF}}\\
    Batch Size & 64 & 256 & 256 & 256 & - \\
    Learning Rate & 5e-4 & 3e-4 & 3e-4 & 3e-4 & - \\
    Discount & 0.99 & 0.99 & 0.99 & 0.99 & - \\
     UTD Ratio & 4 & 4 & 4 & 4 & - \\  
  \midrule
    \multicolumn{1}{@{}l}{\textbf{IWR}}\\
    Learning Rate & 5e-6 & 5e-6 & 1e-4 & 1e-6 & 1e-5\\
    Number of Iterations & 5 & 5 & 5/20 & 5/20 & 6\\
    Episodes Per Iteration & 1 & 3 & 3/1 & 3/1 & 3\\
    Training Epochs per Iteration & 400 & 1000 & 1000/300 & 1000/300 & 500 \\
  \midrule
    \multicolumn{1}{@{}l}{\textbf{Sirius}}\\
    Learning Rate & 5e-6 & 5e-6 & 1e-4 & 1e-6 & 1e-5\\
    Number of Iterations & 5 & 5 & 5/20 & 5/20 & 6\\
    Episodes Per Iteration & 1 & 3 & 3/1 & 3/1 & 3\\
    Training Epochs per Iteration & 400 & 1000 & 1000/300 & 1000/300 & 500 \\
  \midrule
    \multicolumn{1}{@{}l}{\textbf{All Methods}}\\
    Policy Hidden Dims & (64, 64) &(256, 256) & (256, 256) & (256, 256) & (256, 256) \\
  \bottomrule                          
\end{tabular}
}
\end{table}

\end{document}